\documentclass[runningheads]{llncs}
\usepackage[T1]{fontenc}
\usepackage{subcaption}
\usepackage{breakcites}
\usepackage{url}

\usepackage{graphicx}
\begin{document}
\title{On the Role of Activation Functions in EEG-To-Text Decoder}
%
\author{Zenon Lamprou\orcidID{0000-0003-0042-5036} \and
Iakovos Tenedios\orcidID{0009-0004-5727-0455} \and
Yashar Moshfeghi\orcidID{0000-0003-4186-1088}}
\authorrunning{Lamprou et al.}
%
\institute{NeuraSearch Lab, University Of Strathclyde 
\email{\{zenon.lamprou,iakovos.tenedios,yashar.moshfeghi\}@strath.ac.uk}
}

\maketitle              
\begin{abstract}
In recent years, much interdisciplinary research has been conducted exploring potential use cases of neuroscience to advance the field of information retrieval. Initial research concentrated on the use of fMRI data, but fMRI was deemed to be not suitable for real-world applications, and soon, research shifted towards using EEG data. In this paper, we try to improve the original performance of a first attempt at generating text using EEG by focusing on the less explored area of optimising neural network performance. We test a set of different activation functions and compare their performance. Our results show that introducing a higher degree polynomial activation function can enhance model performance without changing the model architecture. We also show that the learnable 3rd-degree activation function performs better on the 1-gram evaluation compared to a 3rd-degree non-learnable function. However, when evaluating the model on 2-grams and above, the polynomial function lacks in performance, whilst the leaky ReLU activation function outperforms the baseline. 

\end{abstract}

\section{Introduction}
In recent years, there have been significant advancements in the study of the human brain. Various researchers have achieved notable success in classifying mental states, ranging from elementary perceptions like pain to more intricate cognitive conditions such as assessing the relevance of stimuli \cite{pinkosova2022revisiting}.
One of the most recent advancements made in the field of neuroscience was the decoding of thoughts and inner speech. Such research was motivated by the pressing challenge of restoring the communication abilities of individuals facing neurological conditions which impaired them. Several researchers were able to decode human thoughts using fMRI \cite{toneva_interpreting_2019} or EEG \cite{wang2022open} data. In the process of doing so, a notion of what we call "brain embeddings" was formed. The models, in their process of learning how to decode human thoughts, built inner representations that map natural language to brain features.

However, using fMRI data is deemed not to be a viable solution for a real-life Brain-to-Text decoder for two reasons. First, it is costly to record fMRI data, and fMRI cannot be used in real-time due to the delay of the BOLD signal that it relies on to identify activation patterns.
Due to these complications, research started moving towards using EEG for real-time Brain-to-Text decoding. With the introduction of ZuCo-1.0 and ZuCo-2.0 \cite{hollenstein_zuco_2020,hollenstein_zuco_2018}, access to a massive dataset that can serve as a training collection was given to the community. 
Previous research has utilised EEG data from ZuCo to either decode human thought or construct general EEG embeddings that can be used in downstream tasks e.g. EEG mental state classification \cite{wang2022open,kostas_bendr_2021,duan_dewave_2023}.
Wang and Ji \cite{wang2022open} attempted first to build an EEG-to-text decoder using EEG data. Even though the results seemed promising, there is a lot of room for improvement to achieve good brain-to-text translation performance.

In this paper, we try to improve the original performance reported by \cite{wang2022open} by delving into the less explored area of optimising neural network performance. We experiment by testing a set of different activation functions and log their performance against the original results. Several works \cite{goyal_learning_2020,bilonoh_tunable_2022} showed that using nonlinear and tunable activation functions can enhance model performance. They also suggested that there is a gap in the knowledge on using such activation functions since all the state-of-the-art models use the most well-known activation functions. However, it is indicated that non-linear and tunable activation functions might fit better to real-life data.
In this paper, we tested a set of different activation functions and compared their performance. Our results show that introducing tunable activation functions can enhance the performance of the model without changing the model architecture. We also show that introducing higher degree polynomial activation functions enhances the performance compared to a linear activation function.
Our main contribution in this paper is to show that it is possible to optimise a Transformer Encoder by using different activation functions either by introducing higher degree polynomial or tunable activation functions.


\section{Related Work}\label{related_work}

\subsubsection{NeuraSearch Research}\label{neurasearch}
In recent years, a lot of interdisciplinary research has been conducted exploring potential use cases of neuroscience to advance the field of information retrieval \cite{Moshfeghi_relevance_2013,mueller2008electrophysiological,muller2015electroencephalography,gwizdka2019introduction}, with such research being named as NeuraSearch \cite{moshfeghi2021neurasearch}. 
Potential NeuraSearch applications have previously been explored in areas such as 
realising the information need of users~\cite{moshfeghi_understanding_2016,moshfeghi2019towards,michalkova_information_2024,2f3779726bd94ea39a8636f049b28d9b}, 
forming search queries \cite{kangassalo2019users}, 
modelling user search process as a transition between different search states~\cite{moshfeghi2018search}, 
evaluating the potential of brain activation for the provision of relevance feedback~\cite{Moshfeghi_relevance_2013,allegretti_when_2015,eugster2016natural,kim2019erp,pinkosova2020cortical,pinkosova2022revisiting,PINKOSOVA2023100295},  
satisfying user information need \cite{paisalnan2021neural,paisalnan2022neural}, and
mental workload detection \cite{McGuire_Ensemble_2023,Kingphai_Channel_2023,kingphai_eeg_2021,Kingphai_Mental_2021}
Many different neuroimaging techniques have been utilised for the purposes of NeuraSearch research, including 
Magnetoencephalography (MEG) \cite{kauppi2015towards}, 
fMRI~\cite{lamprou_role_2022,moshfeghi2013understanding,moshfeghi2019neuropsychological,moshfeghi_understanding_2016,paisalnan2021neural,paisalnan2021towards,paisalnan2022neural}, and 
Electroencephalography (EEG) \cite{allegretti_when_2015,gwizdka2017temporal,jacucci2019integrating,jimenez2018using,kangassalo2019users,kingphai_eeg_2021,kingphai_time_2023,McGuire_Song_2023}. Advantages of some of these techniques and their suitability for the purposes of our research are discussed below.

\subsubsection{EEG-To-Text}\label{eeg_to_text}
Existing research has attempted to decode language in the brain either from text read or speech listened to \cite{herff2015brain,toneva_interpreting_2019,panachakel2021decoding,wang2022open}, which is also the domain that the current research expands upon.
Here, we focus on some research that has assessed the possibility of deriving text from non-invasive brain imaging data \cite{toneva_interpreting_2019,wang2022open}. By utilising fMRI and MEG recordings, Toneva and Wehbe \cite{toneva_interpreting_2019} attempted to improve the natural language processing (NLP) capabilities of various transformer models. They used the brain activation representation of several words from the aforementioned recordings and were able to show that brain activity can be predicted by aligning transformers with the brain activity recordings. They also suggest that such an alignment can lead to increased performance of the transformer model in NLP tasks.

Wang and Ji \cite{wang2022open} used EEG recordings from an open vocabulary dataset which includes tens of thousands of words, and transformer models, to attempt to derive the text that participants were reading from their EEG recordings. They also attempted to enhance transformer EEG-based sentiment classification capabilities through the utilisation of the EEG recordings in the fine-tuning process. They achieved beyond-chance results in decoding participants' EEG data and improving transformer EEG-based sentiment classification performance. One advantage of the data utilised by the latter research is that EEG data has high temporal accuracy, which can allow for real-time decoding of a person’s brain activity, in contrast to the delayed temporal representation of fMRI recordings. Additionally, the acquisition and maintenance costs for EEG devices are much lower compared to fMRI scanners. Therefore, in this research, we would preferably utilise EEG in our experiments due to the enhanced real-life applicability of the data. 

This research will be closely aligned to previous research conducted in decoding EEG-to-text by utilising the same dataset and building upon their methodology \cite{wang2022open,hollenstein_reading_2021}. The aforementioned open vocabulary EEG dataset was collected and presented in two iterations: ZuCo-1.0 and ZuCo-2.0 \cite{hollenstein_reading_2021}. In these datasets, subjects were asked to read sentences with stimuli presentation, being segmented into two modalities: standard and task-specific reading. Specifically, ZuCo-1.0 comprises data from 12 healthy adults, encompassing 21,629 words and 1,107 sentences \cite{hollenstein_zuco_2018}. In contrast, ZuCo-2.0 encapsulates readings from 18 subjects, summing up to 15,138 words in 739 sentences \cite{hollenstein_zuco_2020}. The datasets were constructed by creating eye-tracking features as detailed in the GECO corpus \cite{cop2017presenting}, at both the word and sentence level. Some example features include first fixation duration, which is how long participants fixated on the current word, and total reading time, which includes the total of all fixation duration on the current word. Now, for each eye tracking feature, EEG features, which comprise frequency bands, were extracted, namely theta1, theta2, alpha1, alpha2, beta1, beta2, gamma1 and gamma2. The dataset also includes the raw EEG data for each word and each sentence.

Wang and Ji \cite{wang2022open} utilised a zero-shot pipeline for the classification of EEG to text, with their top-performing pipeline utilising the BART transformer. Their approach is currently yielding a BLEU score of approximately 10\% using constructed word-level EEG features provided in the ZuCo dataset. Even though the original score obtained by the authors was reported to be around 40\% due to an error in the code, which enforced teacher learning. The revised score obtained was approximately 10\% using features and 7\% using the raw EEG recordings. Thus, as this research will closely follow their research, we will also utilise the ZuCo datasets, and we will attempt to enhance the performance of their best-performing pipeline by re-configuring the BART transformer on which it is based. Instead of utilising constructed features, we will be working with the raw EEG data. Such data have been found to perform better in classification tasks when using deep learning or transformer models compared to EEG spectral features \cite{truong2021deep,Siddhad_Gupta_Dogra_Roy_2024}, and abolish the need for pre-processing data reducing required resources.

\subsubsection{Activation Functions}\label{activation_background}

The performance enhancement our research will focus on is the application of several different activation functions, which may improve the performance of the BART transformer \cite{goyal_learning_2020,wang_new_2022,bilonoh_tunable_2022}. Goyal et al. \cite{goyal_learning_2020} highlighted the existing research gap in devising activation functions for neural networks. They introduced an innovative strategy that revolves around constructing learnable activation functions during training. Such functions pertain to a weighted sum of known basis elements which can approximate well the ideal activation function, thereby tailoring to the model's task and data more effectively. 

The incorporation of a novel set of polynomial functions for deep learning was explored by Wang et al. \cite{wang_new_2022}, with their focus being on the domain of precipitation forecasting. They posited that Chebyshev Polynomial-based activation functions paired with normalisation to tackle the data flow explosion issue, which causes them to result in unstable training, might offer a better fit for real-world data, and with appropriate model training, they can boost performance. 

Bilonoh et al. \cite{bilonoh_tunable_2022} proposed another category of activation functions, as they aimed to go beyond the universally used activation functions to speed up convergence and improve accuracy. Their proposed function is termed the tunable activation function, which allows for the modification of its parameters, as well as the between neuron connection weights during the training process. They were found to indeed lead to greater flexibility as, due to their polynomial nature, they can approximate many other types of activation functions. This led to enhanced model performance and more rapid convergence. Thus, in this research, we aim to utilise a variety of the activation functions mentioned above, as it would be interesting to explore whether these activation functions can improve the EEG-to-text capabilities of the BART transformer.


\section{Methodology}\label{methodology}
In order to determine if different activation functions can yield better performance, we decided to re-train the EEG-to-Text decoder proposed by Wang and Ji \cite{wang2022open}. We used the same architecture but implemented different activation functions and ran the experiment using the raw EEG data. We first train the model without any modifications to obtain a baseline, and then we compare our results against that baseline.

\subsection{Data}
For training our model, we utilised the raw EEG data recorded for each word in each sentence. While our model's architecture shares similarities with the one proposed by Wang and Ji \cite{wang2022open}, a key distinction is our use of raw EEG data instead of word-level EEG features. ZuCo dataset also compromises from a variety of tasks. Wang and Ji \cite{wang2022open} showed that by increasing the training sample, the model performance increases. Thus, following their observation, we used Task 1 and Task 2 from ZuCo-1.0 and Task 2 from ZuCo-2.0 for our training.

To accommodate the differences in data formats, we applied transformations to our raw EEG data before fitting it into our model for training. Unlike the word-level features, which are formatted as a 1D series of features, raw EEG data are inherently 2D. The first dimension represents the number of recordings per word, and the second dimension represents the activation levels of each electrode.

To transform the data into the required 1D format, we averaged the activation levels of each electrode across all recordings for a given word. This averaging process yielded a single average activation value per electrode, resulting in a 1D list of features. This transformation allowed us to effectively utilise the raw EEG data within our model, ensuring compatibility and facilitating the training process.

\subsection{Model}

Our model architecture bears a resemblance to the architecture proposed by Wang and Ji \cite{wang2022open}, with a notable difference in the activation functions used for our Transformer Encoder. This architecture is premised on the translation task framework, utilising encoder and decoder neural models. The encoder is composed of a stack of randomly initialised Transformer Encoder layers. These layers aim to extract high-level features from raw EEG, which are then utilised by the decoder. Unlike the original study, we employ different activation functions within these Transformer layers to potentially enhance the feature extraction process. We specifically used 6  Transformer Encoder layers stacked on top of each other with a dimensionality of N and S number of attention heads. For the decoder, we utilise a pre-trained Large Language Model (LLM), specifically BART, which was also used in the original study by Wang and Ji \cite{wang2022open}. BART is particularly well-suited for generation tasks, making it an apt choice for translating EEG-derived features into coherent sentences.

\subsubsection{Model Operation}
Encoder: The randomly initialised stack of Transformer Encoder layers processes the raw EEG data, extracting meaningful features that encapsulate the information within the EEG signal.
Decoder: The BART model takes these high-level features from the encoder and generates the corresponding sentences.
This combination of a customised Transformer Encoder and a robust LLM decoder allows our model to effectively perform the EEG-to-text translation task.
\subsection{Activation Functions}
During our training and evaluation phase, we tested several different activation functions. Below, we give a brief explanation of each activation function. We look for activation functions that outperform ReLU because this is the default activation function used in the Transformer Encoder implementation.

\noindent{\bf Swish.}
Swish activation function \cite{ramachandran_searching_2017} is a smooth, non-monotonic activation function. We chose swish because as a smooth, non-monotonic alternative activation function in deep learning models. Swish has gained popularity after outperforming ReLU on a variety of tasks, such as image classification on ImageNet, and it is reported that it can capture more complex relationships between the data.

\noindent{\bf GELU.}
GELU \cite{hendrycks_gaussian_2023} is a smooth and non-monotonic activation function. GELU weights inputs according to their value and produces richer gradients than ReLU. It outperforms ReLU in a variety of deep learning tasks, particularly when it comes to NLP models like BERT, and that is why it was chosen.

\noindent{\bf ELU.}
ELU \cite{clevert_fast_2016} offers improved performance in various deep-learning applications, such as vanishing gradients and biased activations. The Exponential Linear Unit (ELU) is a smooth, non-monotonic activation function that combines properties of ReLU for positive inputs with a curved exponential saturation for negative inputs, controlled by a parameter $\alpha$.

\noindent{\bf Leaky \& Parametric ReLU.}
We chose Leaky ReLU \cite{maas_rectifier_2013} and Parametric ReLU(PReLU) \cite{he_delving_2015} because they permit non-zero gradients for negative inputs in an effort to solve the dying ReLU problem. While PReLU discovers the ideal slope during training, Leaky ReLU uses a preset slope, which could result in improved performance at the expense of more complexity.

\noindent{\bf Sine.}
As most of the functions mentioned are monotonic, we decided to use a non-monotonic activation function. The sine function and its variations add oscillatory behaviour to neural networks, which may be advantageous for periodic input and tasks requiring non-monotonic modelling skills. However, they are not as popular as ReLU or other activation functions.

\noindent{\bf Chebysev Polynomials.}
Chebyshev polynomials are a significant family of orthogonal polynomials with many analytical features and uses in scientific computing, numerical analysis, and approximation theory. They are defined by a trigonometric cosine formula. As mentioned in Section \ref{related_work}, Chebysev polynomials improved the performance of predicting real-life weather data \cite{wang_new_2022}, so they are deemed to be a good candidate for improving the performance of the model.

\noindent{\bf Learnable Polynomials.}
Learnable polynomials are polynomial equations whose constants are learned during training. In recent research \cite{bilonoh_tunable_2022}, their effectiveness was explored, so this makes them a strong candidate to be included in our experimental set-up.

\subsection{Training process and Evaluation}
Our training process is divided into two identical training runs. In the first run, the learnable parameters of BART are frozen, and we utilise the loss generated from BART to train the parameters of our Transformer Encoder. This loss is the standard loss provided by BART.
During this phase, high-level features are extracted from the Transformer Encoder. These features are then used as input to our BART model. They enable the Transformer Encoder to learn to extract and align EEG signal features effectively with the textual information used by BART.
This initial training run ensures that the Transformer Encoder captures meaningful representations of the EEG data that can be used in the subsequent stages of the model.

The second run follows a similar approach, but this time, we fine-tune BART alongside our Transformer Encoder. This allows the weights of BART to be trained during this phase.
The purpose of conducting two training runs in this manner is twofold: firstly, to train the Transformer Encoder by leveraging the information from BART, enabling it to align EEG information with textual information effectively. Secondly, after this alignment is achieved, we fine-tune BART to perform the EEG-to-text translation as a downstream task.
Once the model training is completed, we use it to generate sentences from BART using beam search. The generated sentences are then evaluated by comparing them to the actual sentences. Metrics such as the BLEU score and Rouge score are computed to assess the quality of the generated text.

We trained the model for each activation function for 10 epochs for each run, for a total of 20 epochs. We used a batch size of 16 and a learning rate of 5e-5 for both runs. To evaluate our model, we concatenated all the data from all the subjects and then performed a random 80\% (train), 10\% (validation), and 10\% (test) split. For both stages of our training, we used a stochastic gradient decent optimiser with a learning rate of $5e-5$.

\subsection{Evaluation Metrics}\label{metrics}
We use two different evaluation metrics for our evaluation process. The first metric was BLEU (Bilingual Evaluation Understudy) \cite{Siddhad_Gupta_Dogra_Roy_2024}.BLEU score is a widely used metric for evaluating the quality of text generated by machine translation models, such as those used in NLP tasks. It measures how closely the generated text matches one or more reference translations. The BLEU score ranges from 0 to 1, where a score closer to 1 indicates a higher quality of translation.

\begin{table}[htbp!]
    \centering
    \caption{BLEU Scores results table}
\begin{tabular}{ |p{0.4\linewidth}||p{0.18\linewidth}|p{0.18\linewidth}| } 
 \hline
 \multicolumn{3}{|c|}{Bleu Scores} \\
 \hline
 Activation Functions & BLEU-Score 1 &BLEU-Score 2\\
\hline
plain & 0.074 &0.0096 \\
\hline
swish& 0.07    &0.0057\\
\hline
gelu& 0.04    &0.008\\
\hline
elu& 0.083    &0.017\\ 
\hline
leaky relu& 0.091    &\textbf{0.028}\\ 
\hline
swish norm first& 0.08   &0.001\\
\hline
parametric relu& 0.075    &0.021\\
\hline
sine& 0.004    &0.0010\\
\hline
chebysev degree 3& 0.087   &0.02\\
\hline
chebysev degree 3 norm first& 0.008    &0.001\\
\hline
chebysev degree 2& 0.095    &0.016\\
\hline
torch poly 2 & 0.083  &0.017\\
\hline
torch poly 3 & \textbf{0.098}   &0.023\\
\hline
negative positive poly & 0.07   &0.015\\
\hline
negative positive poly norm first   & 0.008    &0.001\\
\hline
\end{tabular}
\label{tab:bleu_scores}
\end{table}

BLEU calculates the precision of n-grams (contiguous sequences of n items from a given sample of text) in the generated text compared to the reference text. Commonly, n-grams of sizes 1 to 4 are used.
The second metric was ROUGE (Recall-Oriented Understudy for Gisting Evaluation) \cite{lin-2004-rouge}. The ROUGE score is a set of metrics used to evaluate the quality of summaries and translations generated by NLP models. Unlike BLEU, which focuses on precision, ROUGE primarily emphasises recall and overlap between the generated text and reference texts. ROUGE scores are widely used in summarization tasks and other text-generation evaluations. 

There are different variations of the ROUGE-Score. In this paper, we use ROUGE-N, which measures the n-gram overlap between the generated text and reference text. Common n-grams include uni-gram (ROUGE-1) and bi-gram (ROUGE-2). We also used ROUGE-L, which measures the longest common sub-sequence (LCS) between the generated text and reference text, capturing the order and relationship of words.
\section{Results}\label{results}
To evaluate the performance of our models, we use the BLEU score and Rouge score. These metrics are commonly used to access the machine translation and question-answering tasks, and that is why we adopted them as well. An overview of their implementation can be found in Section \ref{metrics}. We compare our results against the original pipeline \cite{wang2022open}, and in our results tables and figures, we referred to that baseline as "plain". It is worth noting that we combine data from all the subjects even though inter-subject variability and noise are major obstacles in EEG. We did that since in the original research \cite{wang_new_2022}, it was shown that increasing the training data can yield better results. We also wanted to have almost the same parameters as the original research in order for our comparison to be most accurate.

\begin{figure}[htbp!]
    \centering
    \includegraphics[width=\linewidth, height=0.65\linewidth]{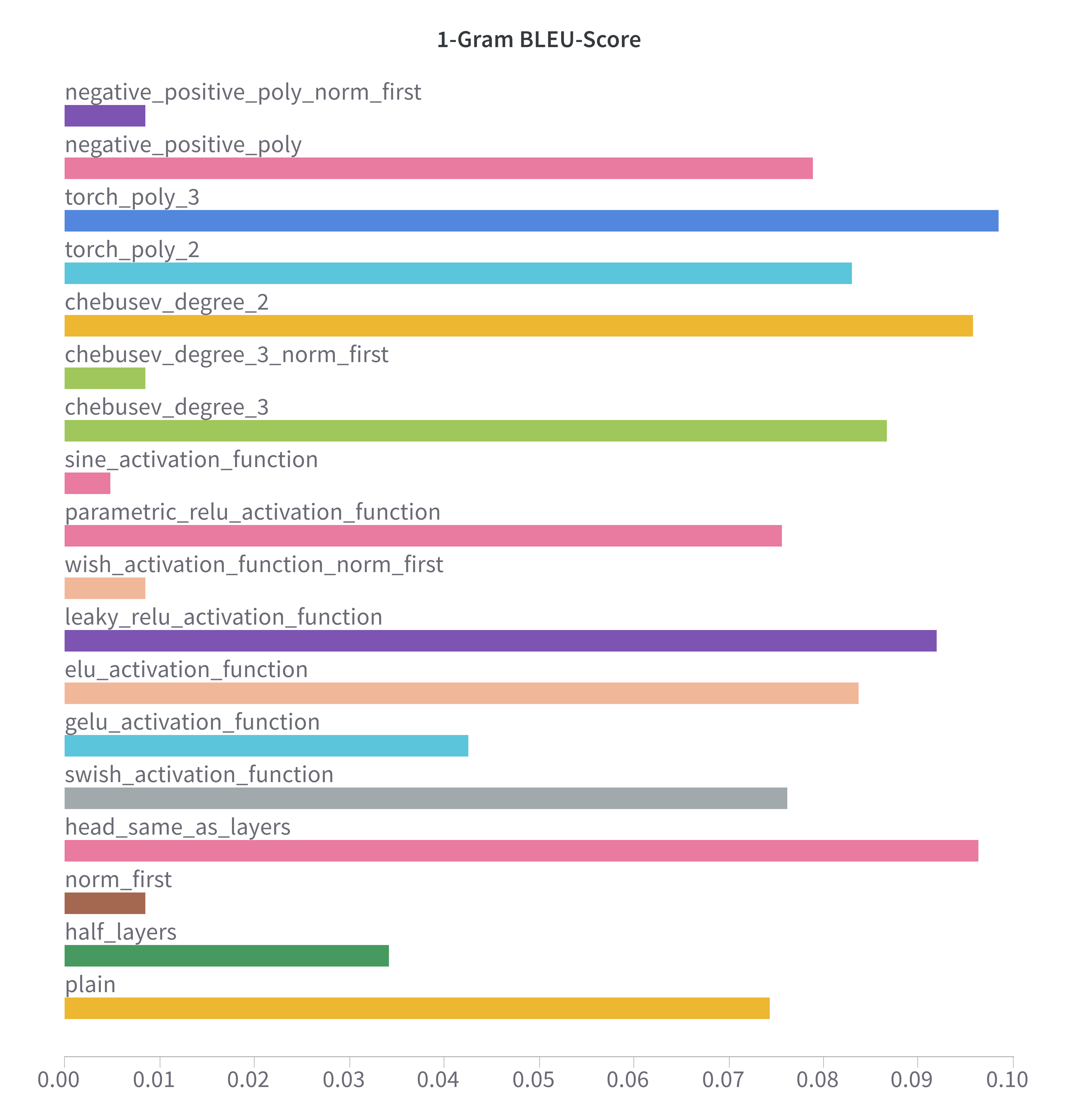}
    \caption{1-Gram BLEU-Score Evaluation}
    \label{fig:1_blue_score}
\end{figure}

\begin{figure}[htbp!]
    \centering
    \includegraphics[width=\linewidth, height=0.65\linewidth]{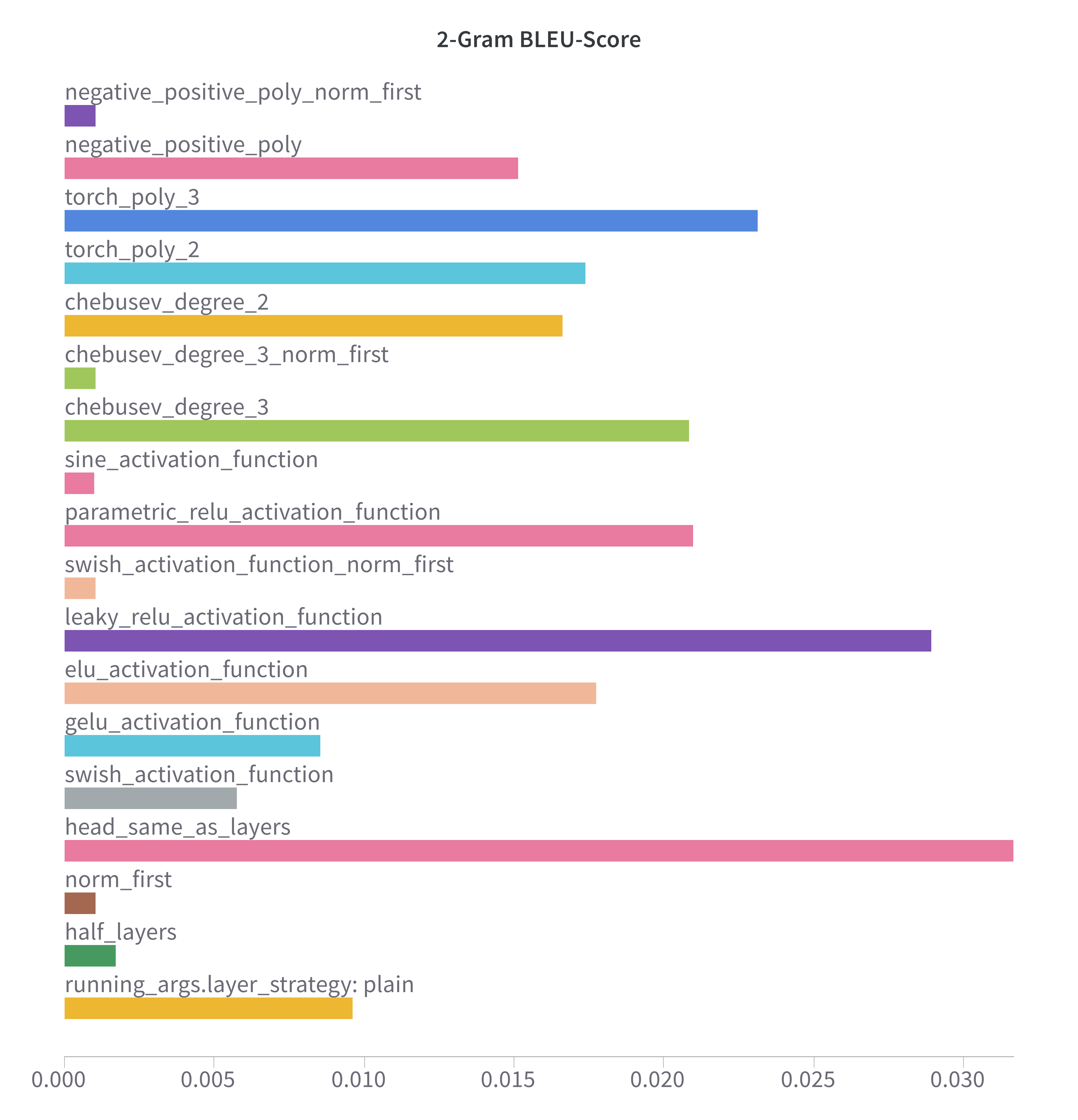}
    \caption{2-Gram BLEU-Score Evaluation}
    \label{fig:2_blue_score}
\end{figure}

Observing Figure \ref{fig:1_blue_score} and \ref{fig:2_blue_score} show that the configurations "torch poly 3", "head same as layers", have the highest BLEU scores, clustered around the 0.09 to 0.10 range. This indicates that these configurations are the most effective in achieving higher translation quality. On the other hand, configurations involving custom activation functions, such as "use custom activation Chebysev degree 3" and "use swish activation function", generally have lower BLEU scores, mostly below 0.08. This wide variance in scores demonstrates the differing levels of effectiveness among the configurations.

\begin{table}[htbp!]
\caption{Rouge Scores results table}
\centering
\begin{tabular}{|p{0.3\linewidth}|p{0.07\linewidth}|p{0.07\linewidth}|p{0.07\linewidth}|p{0.07\linewidth}|p{0.07\linewidth}|p{0.07\linewidth}|p{0.07\linewidth}|p{0.07\linewidth}|p{0.07\linewidth}|}
\hline 
\multicolumn{1}{|c}{Activation Function} & \multicolumn{3}{|c|}{Rouge Score 1} & \multicolumn{3}{c|}{Rouge Score 2} & \multicolumn{3}{c|}{Rouge Score L} \\
\cline{2-10}
\multicolumn{1}{|c|}{} & P & R & F & P & R & F & P & R & F \\
\hline
plain & 0.04 & 0.009 & 0.001 & 0.000 & 0.000 & 0.000 & 0.041 & 0.009 & 0.015 \\
\hline
swish & 0.051 & 0.013 & 0.021 & 0.000 & 0.000 & 0.000  & 0.050 & 0.013 & 0.020\\
\hline
gelu & 0.087 & 0.060 & 0.067 & 0.0026 & 0.0023 & 0.0022 & 0.079 & 0.055 & 0.061 \\
\hline
elu & 0.156 & 0.046 & 0.069 & 0.0037 & 0.0016 & 0.0021 & 0.150 & 0.044 & 0.066 \\
\hline
leaky relu & 0.155 & \textbf{0.115} & \textbf{0.126} & \textbf{0.0142} & \textbf{0.0124}& \textbf{0.0122} & 0.130 & \textbf{0.098} & \textbf{0.106} \\
\hline
swish norm first & 0.018 & 0.013 & 0.015 & 0.0002 & 0.0001 & 0.0002 & 0.018 & 0.013 & 0.014 \\
\hline
parametric relu & 0.158 & 0.114 & 0.114 & 0.0102 & 0.0065 & 0.0074 & 0.131 & 0.083 & 0.095 \\
\hline
sine & 0.009 & 0.008 & 0.008 & 0.0002 & 0.0001 & 0.0001 & 0.009 & 0.008 & 0.008 \\
\hline
chebysev degree 3 & 0.170 & 0.053 & 0.078 & 0.0070 & 0.0028 & 0.0037 & \textbf{0.161} & 0.050 & 0.074 \\
\hline
chebysev degree 3 norm first & 0.018 & 0.013 & 0.015 & 0.0002 & 0.0001 & 0.0001 & 0.018 & 0.013 & 0.014 \\
\hline
chebysev degree 2 & 0.136 & 0.048 & 0.069 & 0.0007 & 0.0002 & 0.0003 & 0.126 & 0.045 & 0.063 \\
\hline
torch poly 2 & 0.106 & 0.041 & 0.057 & 0.0005 & 0.0002& 0.0003 & 0.097 & 0.037 & 0.052 \\
\hline
torch poly 3 & \textbf{0.173} & 0.61 & 0.087 & 0.0055 & 0.0024 & 0.0031 & 0.159 & 0.056 & 0.079 \\
\hline
negative positive poly & 0.142 & 0.044 & 0.065 & 0.0027 & 0.0010 & 0.0014 & 0.140 & 0.043 & 0.063 \\
\hline
negative positive poly norm first & 0.018 & 0.013 & 0.015 & 0.0002 & 0.0001 & 0.0002 & 0.018 & 0.013 & 0.014 \\

\hline
\end{tabular}    
\label{tab:rouge_scores}
\end{table}

After interpreting the results described above, we made several observations. First, the difference in the performance of the Chebysev and Torch polynomials is large even though they use the same 3rd-degree polynomials. We hypothesise that this difference occurs because the torch polynomial has learnable parameters. By adjusting the activation functions’ parameters to be learnable, we can see that the performance of the model can be enhanced. Moreover, the "head is the same as layers" configurations that can happen in combination with a custom activation function. "Head same as layer" initialises a Transformer Encoder that has the same attention head as the number of its layers. By applying that transformation, the learnable activation function outperforms both, which highlights the importance of having a learnable activation function. 

However, the above observation can be seen only if we look at the BLEU score when assessing the performance using a 1-gram evaluation. That implies that our model is good at identifying words in the sentence but is prone to identifying the wrong position of the word. When we look closely at the BLEU score for 2-gram and 3-gram heads, the same layer has the best performance, and the leaky ReLU activation function is the best-performing metric. 

Another observation that can be made is that, by changing the normalisation layer of the encoder to be at the start, the performance drops dramatically to a 10$^{th}$ of the performance regardless of the activation function. As you see in table \ref{tab:bleu_scores}, using the Chebysev polynomial of 3rd degree with the normalisation layer first drops the performance from 0.087 to 0.008. The same observation can be made when using a negative-positive polynomial as well.

To evaluate our models, we also use ROUGE-Score as well. The results can be seen in Table \ref{tab:rouge_scores}. The same observation can be seen for the BLEU-Score. Even though Rouge is a more recall-prone metric for evaluation, we can see that using ROUGE-1 shows that the best performance is achieved when utilising a learnable activation function or using the same head as layers configuration. However, moving up to 2-grams, the leaky ReLU activation function is the best-performing one.

\section{Conclusion \& Future Work}
\label{discussion}
In this paper, we utilise a set of different activation functions and two different transformations to enhance the current performance of an EEG-to-Text decoder model that utilises EEG data. Our results showed that using a different activation function can make the model perform better in real-life data. We also showed the value of using learnable activation functions. Leaky ReLU seemed to be performing better when assessing the 2-gram and above performance in both ROUGE and BLEU scores. The learnable 3rd-degree polynomial function performed better on the 1-gram evaluation. We believe that using different activation functions to train a model and yield better results is an area that has not been vastly explored yet, and our results suggest that it should receive greater attention.

We also believe that even though the results of the EEG-to-Text decoder seemed encouraging, we are still not close to a real-life model that can accurately generate text from raw brain data. However, preliminary research shows that it is possible to generate text from the brain, and this is something we aim to explore in the future. One limitation we identified, and we believe that, if addressed, can improve the model, is the lack of repetition of sentences in the current dataset. Each subject sees each sentence one or two times at most. However, deep learning models rely on repetition in the sample set to learn, so we believe that introducing a repetition to the dataset with more EEG can improve the performance even more. Furthermore, because EEG data tends to be very noise-prone, we hypothesise that a reason that the model doesn't learn better representations is that it doesn't learn the noise because it’s too simple. Another improvement we aim to explore is to introduce a more complex encoder to capture high-level EEG features before providing those features to the large language model. Lastly, we only tested our different activation functions using one seed due to time constraints. Thus, in the future, we aim to test different seeds and determine if the reported performance changes because of the random seed.

\bibliographystyle{splncs04}
\bibliography{biblio}

\begin{thebibliography}{10}
\providecommand{\url}[1]{\texttt{#1}}
\providecommand{\urlprefix}{URL }
\providecommand{\doi}[1]{https://doi.org/#1}

\bibitem{allegretti_when_2015}
Allegretti, M., Moshfeghi, Y., Hadjigeorgieva, M., Pollick, F.E., Jose, J.M.,
  Pasi, G.: When relevance judgement is happening? an {EEG}-based study. In:
  Proceedings of the 38th International {ACM} {SIGIR} Conference on Research
  and Development in Information Retrieval. pp. 719--722. {SIGIR} '15,
  Association for Computing Machinery (2015). \doi{10.1145/2766462.2767811}

\bibitem{bilonoh_tunable_2022}
Bilonoh, B., Bodyanskiy, Y., Kolchygin, B., Mashtalir, S.: Tunable activation
  functions for deep neural networks. In: Babichev, S., Lytvynenko, V. (eds.)
  Lecture Notes in Computational Intelligence and Decision Making. pp.
  624--633. Lecture Notes on Data Engineering and Communications Technologies,
  Springer International Publishing (2022). \doi{10.1007/978-3-030-82014-5_43}

\bibitem{clevert_fast_2016}
Clevert, D.A., Unterthiner, T., Hochreiter, S.: Fast and accurate deep network
  learning by exponential linear units ({ELUs}).
  \doi{10.48550/arXiv.1511.07289}

\bibitem{cop2017presenting}
Cop, U., Dirix, N., Drieghe, D., Duyck, W.: Presenting geco: An eyetracking
  corpus of monolingual and bilingual sentence reading. Behavior research
  methods  \textbf{49},  602--615 (2017)

\bibitem{duan_dewave_2023}
Duan, Y., Zhou, J., Wang, Z., Wang, Y.K., Lin, C.T.: {DeWave}: Discrete {EEG}
  waves encoding for brain dynamics to text translation (2023)

\bibitem{eugster2016natural}
Eugster, M.J., Ruotsalo, T., Spap{\'e}, M.M., Barral, O., Ravaja, N., Jacucci,
  G., Kaski, S.: Natural brain-information interfaces: Recommending information
  by relevance inferred from human brain signals. Scientific reports
  \textbf{6}(1),  38580 (2016)

\bibitem{goyal_learning_2020}
Goyal, M., Goyal, R., Lall, B.: Learning activation functions: A new paradigm
  for understanding neural networks (2020). \doi{10.48550/arXiv.1906.09529}

\bibitem{gwizdka2017temporal}
Gwizdka, J., Hosseini, R., Cole, M., Wang, S.: Temporal dynamics of
  eye-tracking and eeg during reading and relevance decisions. Journal of the
  Association for Information Science and Technology  \textbf{68}(10),
  2299--2312 (2017)

\bibitem{gwizdka2019introduction}
Gwizdka, J., Moshfeghi, Y., Wilson, M.L.: Introduction to the special issue on
  neuro-information science. Journal of the Association for Information Science
  and Technology  \textbf{70},  911--916 (2019). \doi{10.1002/asi.24263}

\bibitem{he_delving_2015}
He, K., Zhang, X., Ren, S., Sun, J.: Delving deep into rectifiers: Surpassing
  human-level performance on {ImageNet} classification.
  \doi{10.48550/arXiv.1502.01852}

\bibitem{hendrycks_gaussian_2023}
Hendrycks, D., Gimpel, K.: Gaussian error linear units ({GELUs}).
  \doi{10.48550/arXiv.1606.08415}

\bibitem{herff2015brain}
Herff, C., Heger, D., De~Pesters, A., Telaar, D., Brunner, P., Schalk, G.,
  Schultz, T.: Brain-to-text: decoding spoken phrases from phone
  representations in the brain. Frontiers in neuroscience  \textbf{9}, ~217
  (2015)

\bibitem{hollenstein_zuco_2018}
Hollenstein, N., Rotsztejn, J., Troendle, M., Pedroni, A., Zhang, C., Langer,
  N.: {ZuCo}, a simultaneous {EEG} and eye-tracking resource for natural
  sentence reading  \textbf{5}(1),  180291 (2018). \doi{10.1038/sdata.2018.291}

\bibitem{hollenstein_zuco_2020}
Hollenstein, N., Troendle, M., Zhang, C., Langer, N.: {ZuCo} 2.0: A dataset of
  physiological recordings during natural reading and annotation. In:
  Proceedings of the Twelfth Language Resources and Evaluation Conference. pp.
  138--146. European Language Resources Association (2020)

\bibitem{hollenstein_reading_2021}
Hollenstein, N., Tröndle, M., Plomecka, M., Kiegeland, S., Özyurt, Y.,
  Jäger, L.A., Langer, N.: Reading task classification using {EEG} and
  eye-tracking data (2021)

\bibitem{jacucci2019integrating}
Jacucci, G., Barral, O., Daee, P., Wenzel, M., Serim, B., Ruotsalo, T.,
  Pluchino, P., Freeman, J., Gamberini, L., Kaski, S., et~al.: Integrating
  neurophysiologic relevance feedback in intent modeling for information
  retrieval. Journal of the Association for Information Science and Technology
  \textbf{70}(9),  917--930 (2019)

\bibitem{jimenez2018using}
Jimenez-Molina, A., Retamal, C., Lira, H.: Using psychophysiological sensors to
  assess mental workload during web browsing. Sensors  \textbf{18}(2), ~458
  (2018)

\bibitem{kangassalo2019users}
Kangassalo, L., Spap{\'e}, M., Jacucci, G., Ruotsalo, T.: Why do users issue
  good queries? neural correlates of term specificity. In: Proceedings of the
  42nd international acm sigir conference on research and development in
  information retrieval. pp. 375--384 (2019)

\bibitem{kauppi2015towards}
Kauppi, J.P., Kandemir, M., Saarinen, V.M., Hirvenkari, L., Parkkonen, L.,
  Klami, A., Hari, R., Kaski, S.: Towards brain-activity-controlled information
  retrieval: Decoding image relevance from meg signals. NeuroImage
  \textbf{112},  288--298 (2015)

\bibitem{kim2019erp}
Kim, H.H., Kim, Y.H.: Erp/mmr algorithm for classifying topic-relevant and
  topic-irrelevant visual shots of documentary videos. Journal of the
  Association for Information Science and Technology  \textbf{70}(9),  931--941
  (2019)

\bibitem{Kingphai_Mental_2021}
Kingphai, K., Moshfeghi, Y.: Mental workload prediction level from eeg signals
  using deep learning models (Sep 2021), the 3rd Neuroergonomics Conference
  2021, NEC21 ; Conference date: 11-09-2021 Through 16-09-2021

\bibitem{kingphai_eeg_2021}
Kingphai, K., Moshfeghi, Y.: On {EEG} preprocessing role in deep learning
  effectiveness for mental workload classification. In: Longo, L., Leva, M.C.
  (eds.) Human Mental Workload: Models and Applications. pp. 81--98.
  Communications in Computer and Information Science, Springer International
  Publishing (2021). \doi{10.1007/978-3-030-91408-0_6}

\bibitem{kingphai_time_2023}
Kingphai, K., Moshfeghi, Y.: On time series cross-validation for deep learning
  classification model of mental workload levels based on eeg signals. In:
  Machine Learning, Optimization, and Data Science. pp. 402--416. Springer
  Nature Switzerland, Cham (2023)

\bibitem{Kingphai_Channel_2023}
Kingphai, K., Moshfeghi, Y.: On channel selection for eeg-based mental workload
  classification. In: Nicosia, G., Ojha, V., La~Malfa, E., La~Malfa, G.,
  Pardalos, P.M., Umeton, R. (eds.) Machine Learning, Optimization, and Data
  Science. pp. 403--417. Springer Nature Switzerland, Cham (2024)

\bibitem{kostas_bendr_2021}
Kostas, D., Aroca-Ouellette, S., Rudzicz, F.: {BENDR}: Using transformers and a
  contrastive self-supervised learning task to learn from massive amounts of
  {EEG} data  \textbf{15} (2021)

\bibitem{lamprou_role_2022}
Lamprou, Z., Pollick, F., Moshfeghi, Y.: Role of punctuation in semantic
  mapping between brain and transformer models. In: Nicosia, G., Ojha, V.,
  La~Malfa, E., La~Malfa, G., Pardalos, P., Di~Fatta, G., Giuffrida, G.,
  Umeton, R. (eds.) Machine Learning, Optimization, and Data Science. pp.
  458--472. Springer Nature Switzerland, Cham (2023)

\bibitem{lin-2004-rouge}
Lin, C.Y.: {ROUGE}: A package for automatic evaluation of summaries. In: Text
  Summarization Branches Out. pp. 74--81. Association for Computational
  Linguistics, Barcelona, Spain (Jul 2004)

\bibitem{maas_rectifier_2013}
Maas, A.L.: Rectifier nonlinearities improve neural network acoustic models

\bibitem{McGuire_Ensemble_2023}
McGuire, N., Moshfeghi, Y.: On ensemble learning for mental workload
  classification. In: Nicosia, G., Ojha, V., La~Malfa, E., La~Malfa, G.,
  Pardalos, P.M., Umeton, R. (eds.) Machine Learning, Optimization, and Data
  Science. pp. 358--372. Springer Nature Switzerland, Cham (2024)

\bibitem{2f3779726bd94ea39a8636f049b28d9b}
McGuire, N., Moshfeghi, Y.: Prediction of the realisation of an information
  need: an eeg study (2024), the 47th International ACM SIGIR Conference on
  Research and Development in Information Retrieval : SIGIR' 24, SIGIR 2024 ;
  Conference date: 14-07-2024 Through 18-07-2024

\bibitem{McGuire_Song_2023}
McGuire, N., Moshfeghi, Y.: What song am i thinking of? In: Nicosia, G., Ojha,
  V., La~Malfa, E., La~Malfa, G., Pardalos, P.M., Umeton, R. (eds.) Machine
  Learning, Optimization, and Data Science. pp. 418--432. Springer Nature
  Switzerland, Cham (2024)

\bibitem{michalkova_information_2024}
Michalkova, D., Rodriguez, M.P., Moshfeghi, Y.: Understanding
  feeling-of-knowing in information search: An eeg study. ACM Trans. Inf. Syst.
   \textbf{42}(3) (jan 2024). \doi{10.1145/3611384}

\bibitem{moshfeghi2021neurasearch}
Moshfeghi, Y.: Neurasearch: Neuroscience and information retrieval. CEUR
  Workshop Proceedings  \textbf{2950},  193--194 (Sep 2021), presented at:
  DESIRES 2021, Design of Experimental Search \& Information REtrieval Systems;
  Proceedings of the Second International Conference on Design of Experimental
  Search \& Information REtrieval Systems, Padova, Italy, September 15-18,
  2021.; 2nd International Conference on Design of Experimental Search and
  Information REtrieval Systems, DESIRES 2021 ; Conference date: 15-09-2021
  Through 18-09-2021

\bibitem{Moshfeghi_relevance_2013}
Moshfeghi, Y., Jose, J.M.: An effective implicit relevance feedback technique
  using affective, physiological and behavioural features. In: Proceedings of
  the 36th International ACM SIGIR Conference on Research and Development in
  Information Retrieval. p. 133–142. SIGIR '13, Association for Computing
  Machinery, New York, NY, USA (2013). \doi{10.1145/2484028.2484074}

\bibitem{moshfeghi2013understanding}
Moshfeghi, Y., Pinto, L.R., Pollick, F.E., Jose, J.M.: Understanding relevance:
  An fmri study. In: Advances in Information Retrieval: 35th European
  Conference on IR Research, ECIR 2013, Moscow, Russia, March 24-27, 2013.
  Proceedings 35. pp. 14--25. Springer (2013)

\bibitem{moshfeghi2018search}
Moshfeghi, Y., Pollick, F.E.: Search process as transitions between neural
  states. In: Proceedings of the 2018 World Wide Web Conference. p.
  1683–1692. WWW '18, International World Wide Web Conferences Steering
  Committee, Republic and Canton of Geneva, CHE (2018).
  \doi{10.1145/3178876.3186080}

\bibitem{moshfeghi2019neuropsychological}
Moshfeghi, Y., Pollick, F.E.: Neuropsychological model of the realization of
  information need. Journal of the Association for Information Science and
  Technology  \textbf{70}(9),  954--967 (2019).
  \doi{https://doi.org/10.1002/asi.24242}

\bibitem{moshfeghi2019towards}
Moshfeghi, Y., Triantafillou, P., Pollick, F.: Towards predicting a realisation
  of an information need based on brain signals. In: The World Wide Web
  Conference. p. 1300–1309. WWW '19, Association for Computing Machinery, New
  York, NY, USA (2019). \doi{10.1145/3308558.3313671}

\bibitem{moshfeghi_understanding_2016}
Moshfeghi, Y., Triantafillou, P., Pollick, F.E.: Understanding information
  need: An {fMRI} study. In: Proceedings of the 39th International {ACM}
  {SIGIR} conference on Research and Development in Information Retrieval. pp.
  335--344. {SIGIR} '16, Association for Computing Machinery (2016).
  \doi{10.1145/2911451.2911534}

\bibitem{mueller2008electrophysiological}
Mueller, V., Brehmer, Y., Von~Oertzen, T., Li, S.C., Lindenberger, U.:
  Electrophysiological correlates of selective attention: a lifespan
  comparison. BMC neuroscience  \textbf{9},  1--21 (2008)

\bibitem{muller2015electroencephalography}
M{\"u}ller-Putz, G.R., Riedl, R., C~Wriessnegger, S., et~al.:
  Electroencephalography (eeg) as a research tool in the information systems
  discipline: Foundations, measurement, and applications. Communications of the
  Association for Information Systems  \textbf{37}(1), ~46 (2015)

\bibitem{paisalnan2021neural}
Paisalnan, S., Moshfeghi, Y., Pollick, F.: Neural correlates of realisation of
  satisfaction in a successful search process. Proceedings of the Association
  for Information Science and Technology  \textbf{58}(1),  282--291 (2021).
  \doi{https://doi.org/10.1002/pra2.456}

\bibitem{paisalnan2021towards}
Paisalnan, S., Pollick, F., Moshfeghi, Y.: Towards understanding neuroscience
  of realisation of information need in light of relevance and satisfaction
  judgement. In: Nicosia, G., Ojha, V., La~Malfa, E., La~Malfa, G., Jansen, G.,
  Pardalos, P.M., Giuffrida, G., Umeton, R. (eds.) Machine Learning,
  Optimization, and Data Science. pp. 41--56. Springer International
  Publishing, Cham (2022)

\bibitem{paisalnan2022neural}
Paisalnan, S., Pollick, F., Moshfeghi, Y.: Neural correlates of satisfaction of
  an information need. In: Nicosia, G., Ojha, V., La~Malfa, E., La~Malfa, G.,
  Pardalos, P., Di~Fatta, G., Giuffrida, G., Umeton, R. (eds.) Machine
  Learning, Optimization, and Data Science. pp. 443--457. Springer Nature
  Switzerland, Cham (2023)

\bibitem{panachakel2021decoding}
Panachakel, J.T., Ramakrishnan, A.G.: Decoding covert speech from eeg-a
  comprehensive review. Frontiers in Neuroscience  \textbf{15},  642251 (2021)

\bibitem{pinkosova2020cortical}
Pinkosova, Z., McGeown, W.J., Moshfeghi, Y.: The cortical activity of graded
  relevance. In: Proceedings of the 43rd International ACM SIGIR Conference on
  Research and Development in Information Retrieval. p. 299–308. SIGIR '20,
  Association for Computing Machinery, New York, NY, USA (2020).
  \doi{10.1145/3397271.3401106}

\bibitem{PINKOSOVA2023100295}
Pinkosova, Z., McGeown, W.J., Moshfeghi, Y.: Moderating effects of
  self-perceived knowledge in a relevance assessment task: An eeg study.
  Computers in Human Behavior Reports  \textbf{11},  100295 (2023).
  \doi{https://doi.org/10.1016/j.chbr.2023.100295}

\bibitem{pinkosova2022revisiting}
Pinkosova, Z., McGeown, W.J., Moshfeghi, Y.: Revisiting neurological aspects of
  relevance: An eeg study. In: Nicosia, G., Ojha, V., La~Malfa, E., La~Malfa,
  G., Pardalos, P., Di~Fatta, G., Giuffrida, G., Umeton, R. (eds.) Machine
  Learning, Optimization, and Data Science. pp. 549--563. Springer Nature
  Switzerland, Cham (2023)

\bibitem{ramachandran_searching_2017}
Ramachandran, P., Zoph, B., Le, Q.V.: Searching for activation functions

\bibitem{Siddhad_Gupta_Dogra_Roy_2024}
Siddhad, G., Gupta, A., Dogra, D.P., Roy, P.P.: Efficacy of transformer
  networks for classification of eeg data. Biomedical Signal Processing and
  Control  \textbf{87},  105488 (Jan 2024). \doi{10.1016/j.bspc.2023.105488}

\bibitem{toneva_interpreting_2019}
Toneva, M., Wehbe, L.: Interpreting and improving natural-language processing
  (in machines) with natural language-processing (in the brain). In: Advances
  in Neural Information Processing Systems. vol.~32. Curran Associates, Inc.
  (2019)

\bibitem{truong2021deep}
Truong, D., Milham, M., Makeig, S., Delorme, A.: Deep convolutional neural
  network applied to electroencephalography: Raw data vs spectral features. In:
  2021 43rd Annual International Conference of the IEEE Engineering in Medicine
  \& Biology Society (EMBC). pp. 1039--1042. IEEE (2021)

\bibitem{wang_new_2022}
Wang, J., Chen, L., Ng, C.W.W.: A new class of polynomial activation functions
  of deep learning for precipitation forecasting. In: Proceedings of the
  Fifteenth {ACM} International Conference on Web Search and Data Mining. pp.
  1025--1035. {WSDM} '22, Association for Computing Machinery (2022).
  \doi{10.1145/3488560.3498448}

\bibitem{wang2022open}
Wang, Z., Ji, H.: Open vocabulary electroencephalography-to-text decoding and
  zero-shot sentiment classification. In: Proceedings of the AAAI Conference on
  Artificial Intelligence. vol.~36, pp. 5350--5358 (2022)

\end{thebibliography}

\appendix
\end{document}